\documentclass[NewProceedings,letterpaper]{ascelike-new}
\usepackage[utf8]{inputenc}
\usepackage[T1]{fontenc}
\usepackage{lmodern}
\usepackage{graphicx}
\usepackage[figurename=Fig.,labelfont=bf,labelsep=period]{caption}
\usepackage{subcaption}
\usepackage{amsmath}
\usepackage{newtxtext,newtxmath}
\usepackage[colorlinks=true,citecolor=black,linkcolor=black]{hyperref}
\begin{document}
\title{Crack Detection of Asphalt Concrete Using Combined Fracture Mechanics and Digital Image Correlation }

\author[1*]{Zehui Zhu, Ph.D.}
\author[2]{Imad L. Al-Qadi, Ph.D.}

\affil[1]{Department of Civil and Environmental Engineering, University of Illinois at Urbana-Champaign. Email: zehuiz2@illinois.edu}
\affil[2]{Department of Civil and Environmental Engineering, University of Illinois at Urbana-Champaign. Email: alqadi@illinois.edu}
\affil[*]{Corresponding Author}

\maketitle

\begin{abstract}
Cracking is a common failure mode in asphalt concrete (AC) pavements. Many tests have been developed to characterize the fracture behavior of AC. Accurate crack detection during testing is crucial to describe AC fracture behavior. This paper proposed a framework to detect surface cracks in AC specimens using two-dimensional digital image correlation (DIC). Two significant drawbacks in previous research in this field were addressed. First, a multi-seed incremental reliability-guided DIC was proposed to solve the decorrelation issue due to large deformation and discontinuities. The method was validated using synthetic deformed images. A correctly implemented analysis could accurately measure strains up to 450\%, even with significant discontinuities (cracks) present in the deformed image. Second, a robust method was developed to detect cracks based on displacement fields. The proposed method uses critical crack tip opening displacement ($\delta_c$) to define the onset of cleavage fracture. The proposed method relies on well-developed fracture mechanics theory. The proposed threshold $\delta_c$ has a physical meaning and can be easily determined from DIC measurement. The method was validated using an extended finite element model. The framework was implemented to measure the crack propagation rate while conducting the Illinois-flexibility index test on two AC mixes. The calculated rates could distinguish mixes based on their cracking potential. The proposed framework could be applied to characterize AC cracking phenomenon, evaluate its fracture properties, assess asphalt mixture testing protocols, and develop theoretical models. 
\end{abstract}

\section{Introduction}
Cracking is a common failure mode in asphalt concrete (AC) pavements. The fracture behavior of AC significantly affects pavement cracking potential. Many tests have been developed to capture the fracture behavior of AC materials. Accurate crack detection during testing is needed to characterize AC fracture behavior.

Contact tools are the commonly used methods. Load, displacement, and/or strain data, recorded during testing are used as indirect measurements of crack development. For example, the load-line displacement, measured by an extensometer mounted vertically at the surface of the specimen, is used to approximate crack propagation in a semi-circular bending test at a low temperature. However, such approximations are insufficient to describe the cracking phenomenon, as a crack tends to choose a path around the aggregate during its growth. 

Computer vision-based crack detection methods have been proposed to detect cracks based on digital images acquired during testing. The most popular algorithms include thresholding, image segmentation, filtering, and blob extraction (\citeNP{hartman2004evaluating}, \citeNP{oliveira2009automatic}, \citeNP{ying2010beamlet},  \citeNP{nisanth2014automated}, \citeNP{zhang2013matched}). However, pixel-level accuracy is difficult to achieve under complex imaging environments. The digital image correlation (DIC) technique has the potential to overcome these challenges.

The DIC is an optical method that can measure displacement and strain. It uses a matching algorithm to establish correspondences between gray value windows extracted from a sequence of images. To detect cracks based on DIC, binary strain contour and bisectional approaches have been proposed (\citeNP{buttlar2014digital}, \citeNP{safavizadeh2017dic}). 

However, two important problems remained to be addressed. 
\begin{itemize}
    \item First, the conventional DIC technique usually fails when a serious decorrelation effect occurs due to large deformation. \cite{pan2012incremental}. For example, the algorithm used by commercial software (e.g., Vic-2D, namely reliability-guided DIC), begins with a manually selected seed point, whose initial guess of the displacement vector is determined using an automatic searching scheme. The full-field displacement is obtained following a calculation path with the highest correlation coefficient.
    Because the shape and position of the target subset change notably under large deformation, it is difficult to obtain an accurate initial guess estimation and select an appropriate shape function or calculation path. Thus, the algorithm is prone to failure. However, accurate DIC measurement is imperative to detect cracks on the surfaces of AC specimens.
    \item Second, existing approaches relied on empirical thresholds or unproven assumptions. For example, \citeN{safavizadeh2017dic} assumed a $e_{xx}$ threshold of 9,000 $\mu\epsilon$ and a $e_{yy}$ threshold of 6,000 $\mu\epsilon$ to identify vertical and interfacial cracks. \citeN{buttlar2014digital} assumed that the deviation point of a relative displacement and number of cycles curve was the failure point.
\end{itemize}

This paper proposed a method to accurately measure large deformation at the post-peak-load stage of AC testing. A synthetic dataset was developed to validate the proposed method. In addition, a robust approach, which relies on the well-developed crack tip opening displacement (CTOD) concept, was developed to detect cracks with DIC. An extended finite element model (XFEM) was created to validate the proposed approach. 

\section{Two-Dimensional Digital Image Correlation Principle}
\begin{figure}[h!]
    \centering
    \includegraphics[trim=0 0 0 0,clip,width=0.55\textwidth]{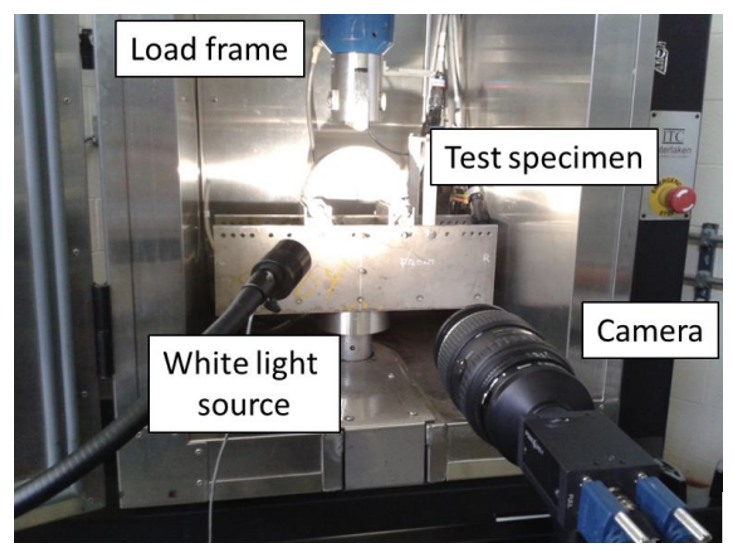}
    \caption{2-D DIC setup.}
    \label{fig:dic_set_up}
\end{figure}

The DIC is a non-contact, full-field displacement/strain measurement technique. A commonly used 2-D DIC system consists of a camera, a lighting system, a computer, and a post-processing program (Fig.\ref{fig:dic_set_up}). 

\subsection{Digital Image Correlation Fundamentals}
\begin{figure}[ht!]
    \centering
    \includegraphics[trim=0 0 0 0,clip,width=0.95\textwidth]{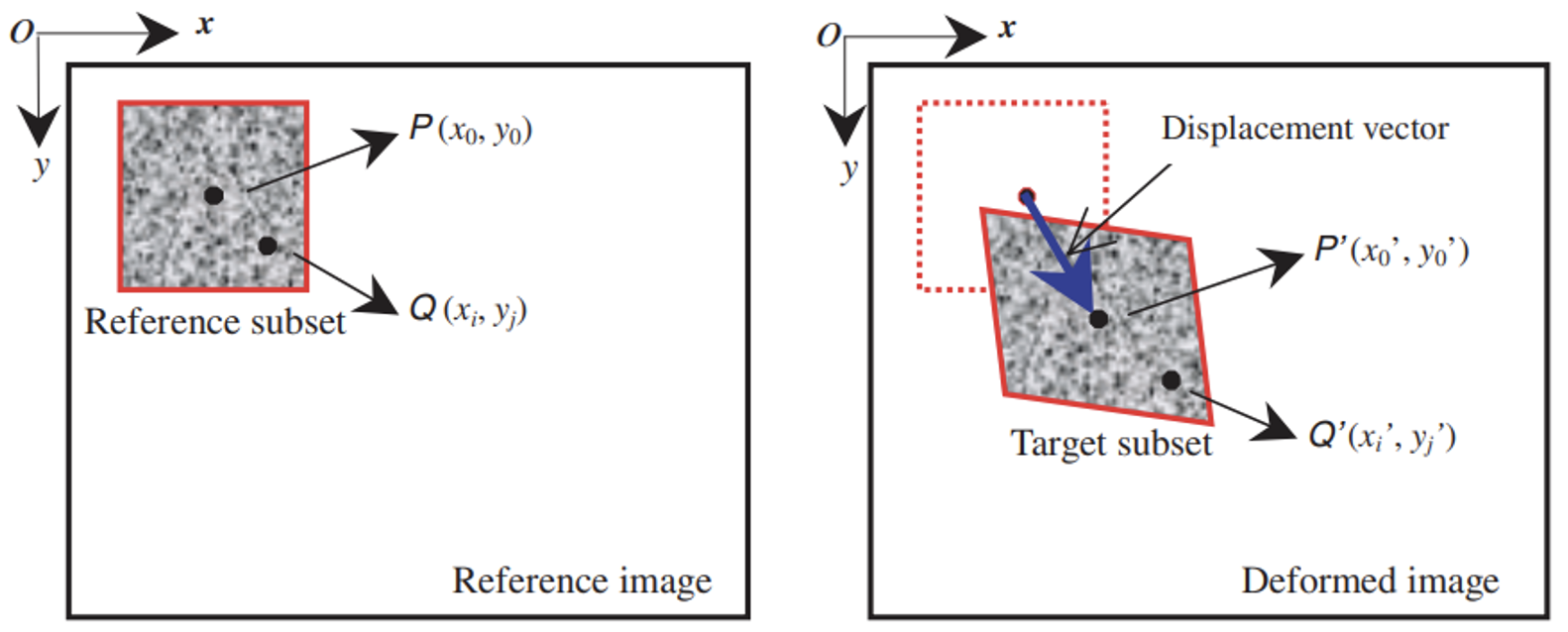}
    \caption{Area-based matching.}
    \label{fig:abm}
\end{figure}

The DIC works by tracking pixels in a sequence of images. It is achieved using area-based matching, which extracts gray value correspondences based on their similarities. As shown in Fig.\ref{fig:abm}, to compute the displacements of point $P$, a square subset of pixels $((2M+1) \times (2M+1))$ from the reference image is chosen to match the deformed image. A region of interest (ROI) must be defined in the reference image and divided into evenly spaced grids. The displacements are computed at each grid to obtain the displacement field. 

The matching is accomplished by minimizing the zero-normalized sum of squared difference (ZNSSD) cost function (Eq.\ref{eqn:cost_fn}), which is insensitive to offset and scale in lighting.

\begin{equation}
    \sum_{i=-M}^{M} \sum_{j=-M}^{M} [\frac{f(x_i,y_j)-f_m}{\sqrt{\sum_{i=-M}^{M} \sum_{j=-M}^{M} [f(x_i,y_j)-f_m]^2}} - \frac{g(x_i',y_j')-g_m}{\sqrt{\sum_{i=-M}^{M} \sum_{j=-M}^{M} [g(x_i',y_j')-g_m]^2}}]^2
\label{eqn:cost_fn}
\end{equation}

where $f(x_i,y_j)$ is gray value at $(x_i, y_j)$ in the reference subset. $g(x_i',y_j')$ is gray value at $(x_i', y_j')$ in the deformed subset. $f_m$ and $g_m$ are mean gray values of the target subset in reference and deformed images, respectively. A close to zero ZNSSD cost, or correlation coefficient, indicates a good match. It is worth mentioning that another commonly used cost, zero-mean normalized cross-correlation (ZNCC), is directly related to ZNSSD. Eq.\ref{eqn:cost_relation} suggests that $C_{ZNCC}=1$ implies a perfect match, while $C_{ZNCC}=0$ denotes no correlation.

\begin{equation}
    C_{ZNCC} =\frac{\sum_{i=-M}^{M} \sum_{j=-M}^{M}[f(x_i,y_j)-f_m]\times[g(x_i',y_j')-g_m]}{\sqrt{\sum_{i=-M}^{M} \sum_{j=-M}^{M} [f(x_i,y_j)-f_m]^2}\sqrt{\sum_{i=-M}^{M} \sum_{j=-M}^{M} [g(x_i',y_j')-g_m]^2}}= 1-0.5C_{ZNSSD}
\label{eqn:cost_relation}
\end{equation}

In Eq.\ref{eqn:cost_fn}, the reference point $(x_i,y_j)$ is mapped to the deformed point $g(x_i',y_j')$ according to a displacement mapping function. First-order (Eq.\ref{eqn:displ_first_fn}) and second-order (Eq.\ref{eqn:displ_second_fn}) functions are commonly used. The latter could approximate more complicated displacement than the former. 

\begin{equation}
    \begin{bmatrix} x_i' \\ y_j' \end{bmatrix} = \begin{bmatrix} x_0 \\ y_0 \end{bmatrix}+\begin{bmatrix} 1+u_x & u_y & u \\ v_x & 1+v_y &v \end{bmatrix} \begin{bmatrix} \Delta x \\\Delta y \\ 1 \end{bmatrix}
\label{eqn:displ_first_fn}
\end{equation}

\begin{equation}
    \begin{bmatrix} x_i' \\ y_j' \end{bmatrix} = \begin{bmatrix} x_0 \\ y_0 \end{bmatrix}+\begin{bmatrix} 1+u_x & u_y & \frac{1}{2}u_{xx} & \frac{1}{2}u_{yy}& u_{xy}& u\\ v_x & 1+v_y & \frac{1}{2}v_{xx} & \frac{1}{2}v_{yy}& v_{xy}& v \end{bmatrix} \begin{bmatrix} \Delta x \\\Delta y \\ \Delta x^2 \\\Delta y^2 \\\Delta x \Delta y \\ 1 \end{bmatrix}
\label{eqn:displ_second_fn}
\end{equation}

where $u$ and $v$ are displacement components for the subset center $(x_0,y_0)$ in the $x$ and $y$ directions, respectively; $\Delta x=x_i-x_0$;  $\Delta y=y_j-y_0$; $u_x$, $u_y$, $v_x$, and $v_y$ are first-order displacement gradients components; $u_{xx}$, $u_{yy}$, $u_{xy}$, $v_{xx}$, $v_{yy}$, and $v_{xy}$ are second-order displacement gradients components. $\mathbf{p}$ is used to denote the desired displacement vector in this paper, with six or twelve unknown parameters.

With the definitions above, it is clear that calculating $\mathbf{p}$ is an optimization problem for a user-defined cost function like Eq.\ref{eqn:cost_fn} and Eq.\ref{eqn:cost_relation}. The Newton–Raphson (NR) iteration method is used in DIC for optimization (Eq.\ref{eqn:NR}).

\begin{equation}
    \mathbf{p} = \mathbf{p}_0-\frac{\nabla C(\mathbf{p}_0)}{\nabla \nabla C(\mathbf{p}_0)}
\label{eqn:NR}
\end{equation}

where $\mathbf{p}_0$ is an initial guess of the displacement vector; $\mathbf{p}$ is next iterative solution; $\nabla C(\mathbf{p}_0)$ is first-order derivatives of the cost function; and $\nabla \nabla C(\mathbf{p}_0)$ is Hessian matrix \cite{pan2009two}. 

\subsection{Reliability-Guided Digital Image Correlation}
The above section only described the procedure to calculate $\mathbf{p}$ for a single point. The RG-DIC method is commonly used to obtain full-field displacement. It is adopted by commercial software such as Vic-2D and open-source software like Ncorr (\citeNP{pan2009reliability}, \citeNP{blaber2015ncorr}). 

The algorithm starts by obtaining an initial guess of the displacement vector for a user-defined seed point. Normalized cross-correlation or scale-invariant feature transform could be used to search for a reasonable initial guess. Then, $\mathbf{p}_{seed}$ and its corresponding correlation coefficient are computed. Next, the four neighboring points of the seed point are calculated using $\mathbf{p}_{seed}$ as their initial guess. Their correlation coefficients are inserted into a priority queue. Next, the first point in the queue with the highest correlation is removed, and its $\mathbf{p}$ is used as the initial guess to calculate displacements of the four neighboring points if they have not been computed yet. The above step is repeated until the priority queue is empty, implying all points in the ROI have been calculated.

The RG-DIC method is insensitive to small discontinuities in images because the correlation analysis is always performed along with the points with the highest correlation. Such a feature makes it robust in analyzing images of AC specimen surfaces, where irregularities are unavoidably present due to air voids and varying aggregate orientation.

\section{Research Objective}
This paper addresses the following problems in using 2D-DIC to detect cracks on an AC specimen surface.
\begin{itemize}
    \item Conventional 2-D DIC technique usually fails when a profound decorrelation effect occurs due to large deformation, frequently in analyzing deformed images acquired at the post-peak-load stage in AC fracture tests.
    \item There is no validated and generalized method to detect cracks based on displacement or strain field. Existing algorithms rely on either empirical strain thresholds or unproven failure point assumptions.
\end{itemize}

A method that accurately measures large deformation was proposed to achieve the goal. A synthetic dataset was developed to validate the proposed method. Next, a robust approach that relies on the CTOD concept was developed to detect cracks with DIC. The XFEM was used to validate the proposed approach.

\section{Large Deformation Measurement Using 2D-DIC}
\subsection{Multi-Seed Analysis}
\begin{figure}[ht!]
    \centering
    \includegraphics[trim=0 0 0 0,clip,width=0.95\textwidth]{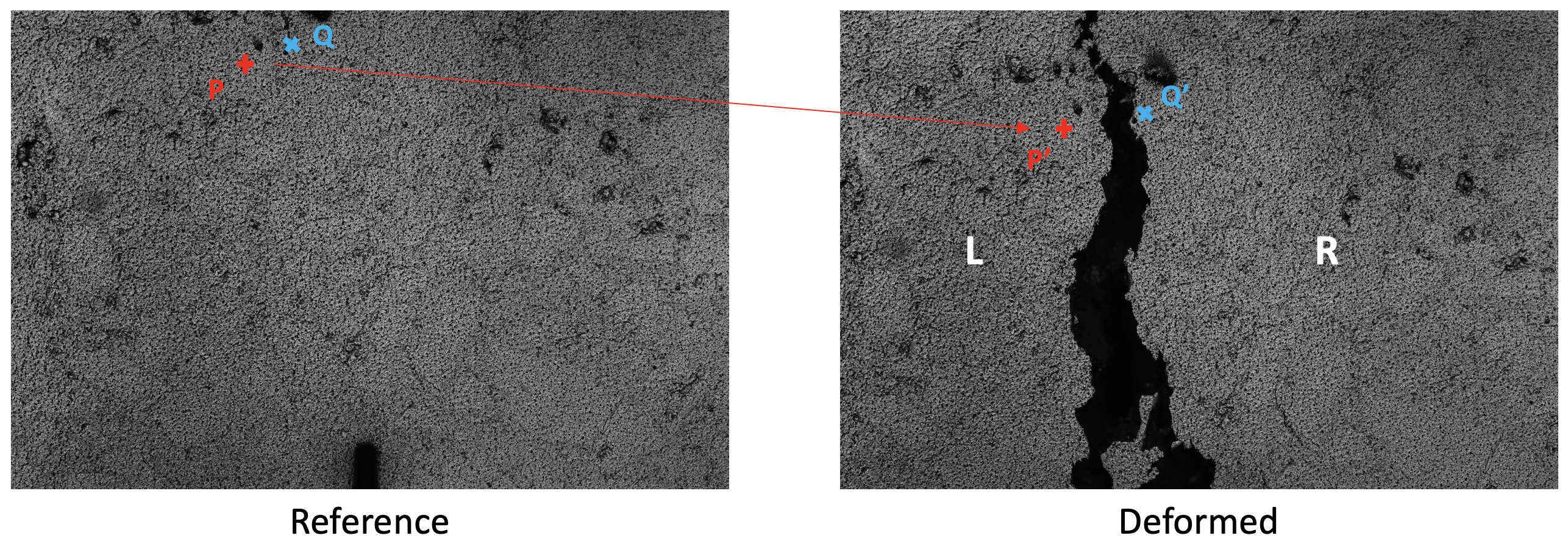}
    \caption{Wide crack under large deformation.}
    \label{fig:multiseed}
\end{figure}

The conventional RG-DIC algorithm typically starts with one user-defined seed point. However, it may fail when a large crack presents in the deformed image. For example, as shown in Fig.\ref{fig:multiseed}, a reference and a deformed image were collected while conducting the Illinois-flexibility index test (I-FIT) on an Illinois N70 AC mix at 25\textdegree{C} with a loading rate of 50 mm/min, in accordance with AASHTO T393. If the seed is placed at point $P$ in partition $L$, the algorithm fails in calculating displacement vectors of points in partition $R$. In RG-DIC, the displacement vector of one of its four neighboring points is used as an initial guess to calculate the displacement vector of a correlation point. As would be expected, when calculating the displacement vector of point $Q$ in partition $R$, the initial guess, which comes from a point in partition $L$, would be far from the ground truth, resulting in correlation failure.

A straightforward solution is a multi-seed analysis. In the example above, if a second seed is placed in partition $R$, the initial guess of the displacement vector of point $Q$ will be closer to the ground truth, and a reliable correlation would be expected. 

\subsection{Incremental Correlation}
Large deformations can cause severe decorrelation such that even the multi-seed RG-DIC technique fails. It frequently happens in analyzing deformed images acquired at the post-peak-load stage of AC testing, where the crack propagates. The incremental correlation was proposed to measure large deformation to solve this problem accurately. Given a series of images, if the $i$th deformed image severely decorrelates with the reference image, an intermediate deformed image ($j$) will be used as an updated reference image. The desired displacement vector of point $(x,y)$ in image $i$ can be calculated using Eq.\ref{eqn:incremental}.

\begin{equation}
    \mathbf{d}_i(x,y) = \mathbf{d}_j(x,y)+	\bigtriangleup \mathbf{d}_{i,j}(x,y)
\label{eqn:incremental}
\end{equation}

where $\mathbf{d}_i(x,y)$ refers to the displacement vector of point $(x,y)$ in the $i$th image with respect to the reference image; $\mathbf{d}_j(x,y)$ is displacement vector of point $(x,y)$ in the $j$th deformed image (updated reference image) with respect to the reference image; $\bigtriangleup \mathbf{d}_{i,j}(x,y)$ is incremental displacement vector at point $(x,y)$ between the $i$th and the updated reference image. 

\subsection{Experiments}
\subsubsection{Synthetic Image Dataset}
\begin{figure}[ht!]
    \centering
    \includegraphics[trim=0 0 0 0,clip,width=0.95\textwidth]{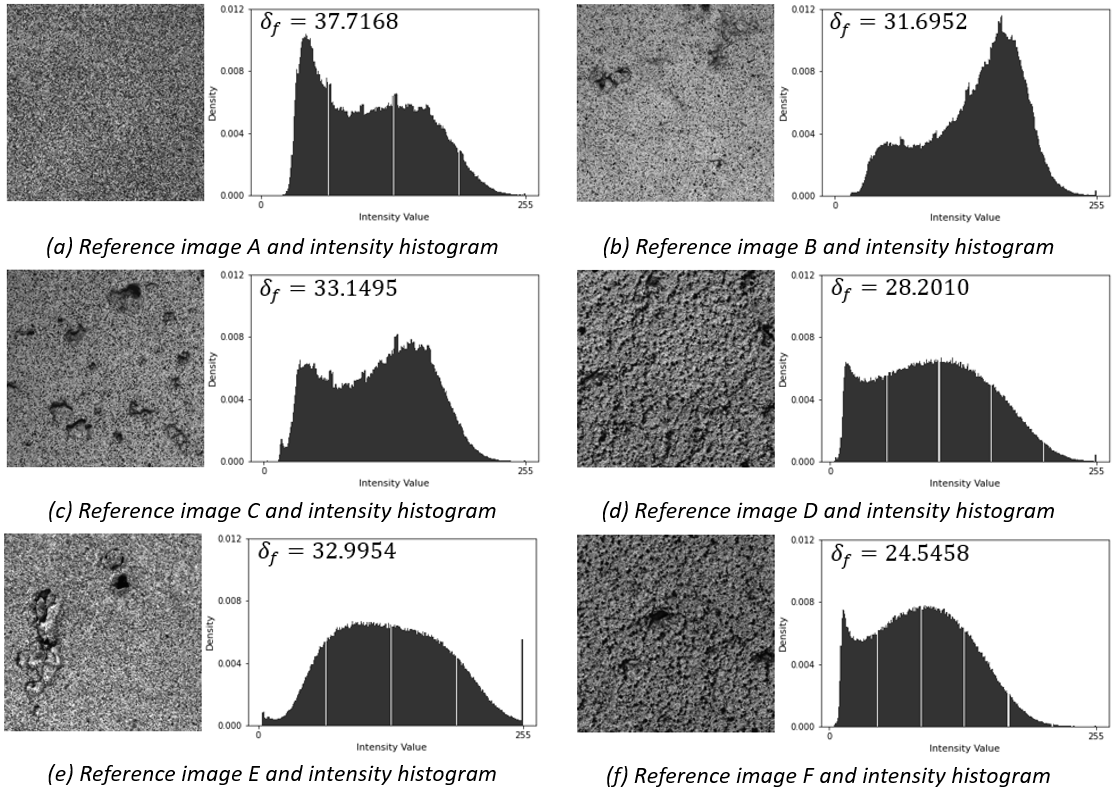}
    \caption{Speckle patterns and intensity histograms of reference images.}
    \label{fig:reference}
\end{figure}

Synthetic image pairs were created to evaluate the performance of multi-seed analysis and incremental correlation for large deformation measurement. Six static images were collected on different AC specimens with varying speckle patterns. Fig.\ref{fig:reference} shows the reference images and their corresponding intensity histograms. Frames A, B, and C were collected using a couple-charged device (CCD) camera with a resolution of $2048 \times 2048$. A random black pattern was applied on top of the white paint. The random black pattern was applied using a spray can of paint. Frames D, E, and F were collected using a CCD camera with a resolution of $6576 \times 4384$. The speckles were sprayed using a fine airbrush. For experimental purposes, all images were cropped to have a dimension of $1024 \times 1024$. The mean intensity gradient parameter was calculated for each image using Eq.\ref{eqn:mig} to assess the quality of the entire speckle pattern \cite{pan2010mean}. Fig.\ref{fig:reference} shows that all speckle patterns have good quality because of their large enough mean intensity gradient. More importantly, the mean intensity gradient was later used to select the appropriate subset size.

\begin{equation}
\delta_f = \frac{\sum_{i=1}^W \sum_{j=1}^H |\nabla f(\mathbf{x}_{ij})|}{W \times H}
\label{eqn:mig}
\end{equation}

where $W$ and $H$ are image width and height in pixels, respectively; $|\nabla f(\mathbf{x}_{ij})|=\sqrt{f_x^2(\mathbf{x}_{ij})+f_y^2(\mathbf{x}_{ij})}$; $f_x(\mathbf{x}_{ij})$ and $f_y(\mathbf{x}_{ij})$ are the intensity derivatives at pixel ${x}_{ij}$ at the $x$- and $y$-direction, respectively. A Prewitt kernel was used to compute intensity derivatives.

\begin{figure}[ht!]
    \centering
    \includegraphics[trim=0 0 0 0,clip,width=0.5\textwidth]{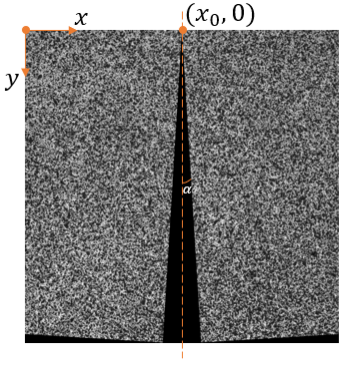}
    \caption{Displacement fields of synthetic deformed images.}
    \label{fig:displ}
\end{figure}

\begin{figure}[ht!]
    \centering
    \includegraphics[trim=0 0 0 0,clip,width=0.95\textwidth]{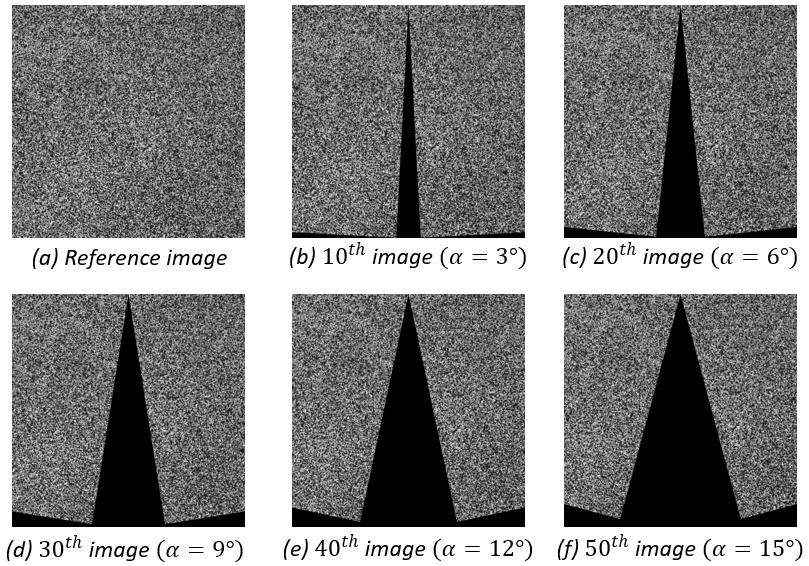}
    \caption{Reference frame A and its corresponding deformed images.}
    \label{fig:img_example}
\end{figure}

One type of displacement field was used to generate deformed images, as shown in Fig.\ref{fig:displ}. It involves rotation with respect to point $(x_0,0)$ only. The displacement vector $(u_x, u_y)$ at each pixel can be calculated using Eq.\ref{eqn:displ}. It should be noted that counterclockwise is defined as the positive rotation direction in this paper. Fifty deformed images were generated for each series of static images with $\alpha = {0.3^\circ, 0.6^\circ, \dots, 15^\circ}$. Fig.\ref{fig:img_example} shows the $10^{th}$, $20^{th}$, $30^{th}$, $40^{th}$, and $50^{th}$ deformed image of reference frame A. Synthetic image series was not simulating a real crack propagation.

\begin{equation}
\begin{cases}
    u_x = \sin (\alpha+\tan^{-1} (\frac{x-x_0}{y}))\sqrt{(x-x_0)^2+y^2} + x_0 - x \\
    u_y = \cos (\alpha+\tan^{-1} (\frac{x-x_0}{y}))\sqrt{(x-x_0)^2+y^2} - y
\end{cases}
\label{eqn:displ}
\end{equation}

\subsubsection{Results}
\begin{figure}[ht!]
    \centering
    \includegraphics[trim=0 0 0 0,clip,width=1\textwidth]{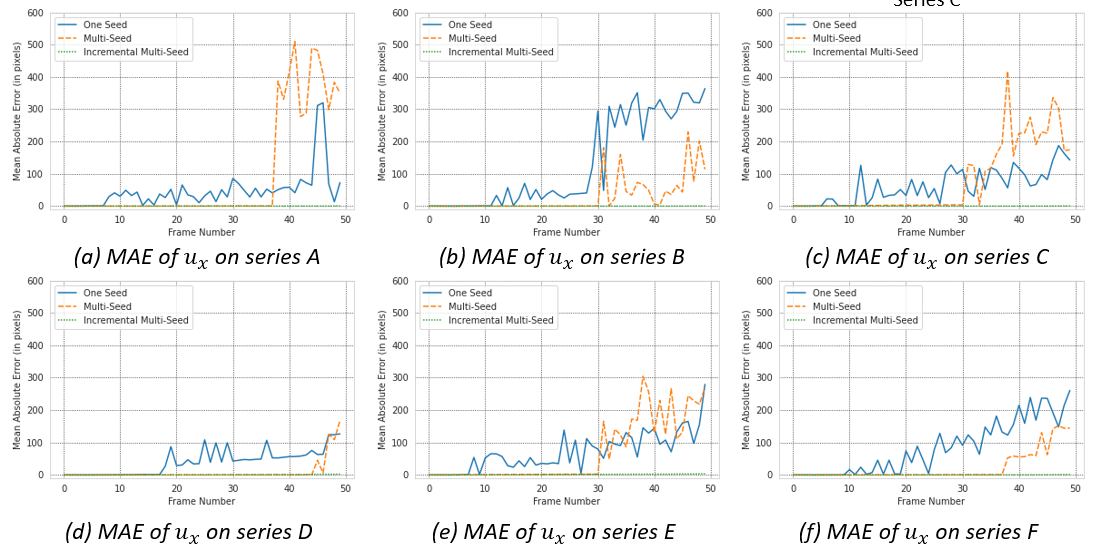}
    \caption{Mean absolute error of DIC analysis.}
    \label{fig:result1}
\end{figure}

An ROI with a dimension of $640\times 824$ was placed at the image center. Three analyses were conducted on each series of images:
\begin{itemize}
    \item RG-DIC analysis with one seed placed in the left partition (referred to as \emph{`One Seed'} in Fig.\ref{fig:result1}).
    \item Multi-seed RG-DIC analysis with one seed in the left partition and one seed in the right partition (referred to as \emph{`Multi-Seed'} in Fig.\ref{fig:result1}).
    \item Multi-seed incremental RG-DIC analysis with one seed in the left partition and one seed in the right partition (referred to as \emph{`Incremental Multi-Seed'} in Fig.\ref{fig:result1}).
\end{itemize}

The mean absolute error (MAE) of $u_x$ and $u_y$ were used as the evaluation criteria. 

\begin{equation}
\begin{gathered}
    \textnormal{MAE}_x = \frac{\sum_{i=1}^{W} \sum_{j=1}^{H} |u_{x_{ij}}'-u_{x_{ij}}|}{W\times H} \\
    \textnormal{MAE}_y = \frac{\sum_{i=1}^{W} \sum_{j=1}^{H} |u_{y_{ij}}'-u_{y_{ij}}|}{W\times H}
\end{gathered}
\label{eqn:mae}
\end{equation}

where $W$ and $H$ are number of correlation points in $x$- and $y$-direction, respectively; $u_{x_{ij}}'$ and $u_{y_{ij}}'$ are calculated displacement in $x$- and $y$-direction, respectively; $u_{x_{ij}}$ and $u_{y_{ij}}$ are ground-truth displacement in $x$- and $y$-directions obtained from Eq.\ref{eqn:displ}, respectively.

Fig.\ref{fig:result1} shows the mean absolute error of $u_x$ on the six series of deformed images. The findings are summarized below:
\begin{itemize}
    \item One seed RG-DIC analysis accurately measured relatively small displacement with small discontinuities (cracks) present in the deformed images. However, it failed when the discontinuities (cracks) became more prominent. Multi-seed RG-DIC analysis accurately measured displacement with relatively more significant discontinuities (cracks). Seeds must be placed on both sides of the crack plane.
    \item Multi-seed incremental RG-DIC analysis could accurately measure the displacement field under large deformation with significant discontinuities (cracks) present in the deformed images. A correctly implemented multi-seed incremental RG-DIC analysis could consistently achieve high accuracy, at least up to a strain level of 450\%, even with significant discontinuities (cracks) present in the deformed image. Moreover, a few irregularities and holes commonly seen on an AC specimen surface have no significant impact on the measurement accuracy.
\end{itemize}

\section{Crack Detection Based on Displacement Field}
\subsection{Method Description}
Surface cracks are defined as displacement field discontinuities. Although the strain field obtained from DIC analysis could highlight the zones that contain discontinuities, the strains across strong discontinuities are theoretically infinite, which induces severe decorrelation in DIC analysis \cite{nguyen2011fracture}. Hence, using DIC measured strain field to detect cracks often fails when the cracks are large. Moreover, strain is not a material property. The strain threshold needs to be adjusted when the testing condition changes. Hence, such a threshold is difficult to obtain. Instead, the method proposed in this paper uses the displacement field to detect surface cracks on AC specimens.

Given displacement fields ($\mathbf{u}, \mathbf{v}$) that consist $N \times M$ square microns in the ROI, the relative displacement ($\mathbf{u}_x$, $\mathbf{v}_y$) between neighboring correlation points can be calculated by filtering $\mathbf{u}$ and $\mathbf{v}$ by a $[-1,1]$ kernel and a $[-1,1]^T$ kernel, respectively. For AC fracture tests where the cracks propagate vertically (in $y-$direction), a $u_{x_i}$ that is greater or equal to a critical CTOD value ($\delta_c$) means the onset of cleavage fracture. Crack edges could be labeled by locating the corresponding correlation points in the deformed image. Similarly, for AC fracture tests where the cracks propagate horizontally (in $x-$direction), a $v_{y_j} \geq \delta_c$ indicates discontinuities. The $\delta_c$ for a specimen configuration could be determined following the procedure discussed in the following section.

The proposed method relies on well-developed fracture mechanics theory. The proposed threshold $\delta_c$ has a physical meaning and can be easily determined from DIC measured displacement field. The concept of CTOD and the determination of $\delta_c$ are discussed in the following section.

\subsection{Crack Tip Opening Displacement}
Crack tip opening displacement is a fracture parameter that can characterize a crack for linear elastic fracture mechanics (LEFM) and elastic-plastic fracture mechanics (EPFM) \cite{kumar2009elements}. The CTOD criterion assumes that fracture occurs when CTOD exceeds $\delta_c$, associated with cleavage fracture onset under plane strain conditions \cite{zhu2012review}. The CTOD criterion has been successfully implemented to define AC specimen crack initiation in the three-point bending (3PB) test and disk-shaped compact tension (DCT) test (\citeNP{song2008delta25}, \citeNP{chen2014crack}). 

\begin{figure}[ht!]
    \centering
    \includegraphics[trim=0 0 0 0,clip,width=1\textwidth]{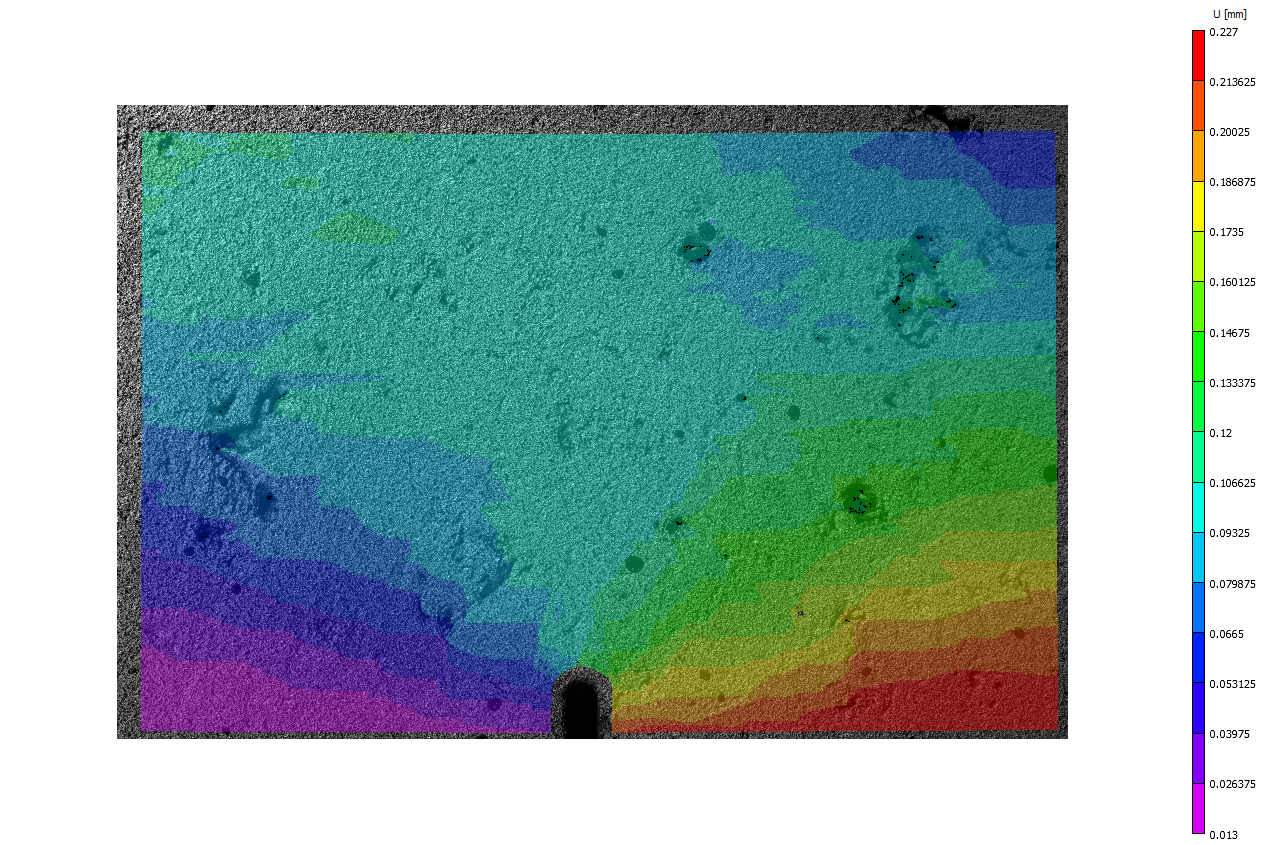}
    \caption{Displacement field $\mathbf{u}$ measured with DIC at the pre-peak load stage in an I-FIT.}
    \label{fig:u}
\end{figure}

\begin{figure}[ht!]
    \centering
    \includegraphics[trim=0 0 0 0,clip,width=1\textwidth]{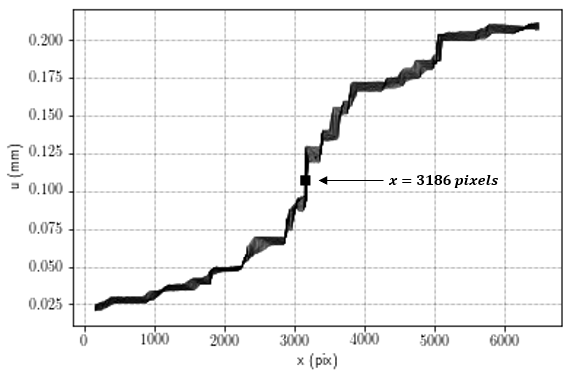}
    \caption{A set of profiles of horizontal displacement perpendicular to the crack plane.}
    \label{fig:tip}
\end{figure}

\citeN{vasco2019characterisation} proposed a method to measure CTOD using DIC. The concept was applied to images acquired while conducting an I-FIT on an Illinois N90 AC mix. First, the crack tip location is determined as it significantly impacts the CTOD measurement accuracy. For AC fracture tests where the cracks propagate vertically (in $y-$direction), given DIC-measured displacement field $\mathbf{u}$ (as shown in Fig.\ref{fig:u}), the $x-$coordinate is found by plotting a set of profiles of horizontal displacement perpendicular to the crack plane (as shown in Fig.\ref{fig:tip}). By locating the intersection point of all profiles, the $x-$coordinate of the crack tip can be determined. The $y-$coordinate could be found from one of the profiles using the $x-$coordinate and its corresponding displacement $u$. For AC fracture tests where the cracks propagate horizontally (in $x-$direction), the $y-$coordinate of the crack tip is determined first, given the displacement field $\mathbf{v}$. The $x-$coordinate could be found subsequently.  

\begin{figure}[ht!]
    \centering
    \includegraphics[trim=0 0 0 0,clip,width=1\textwidth]{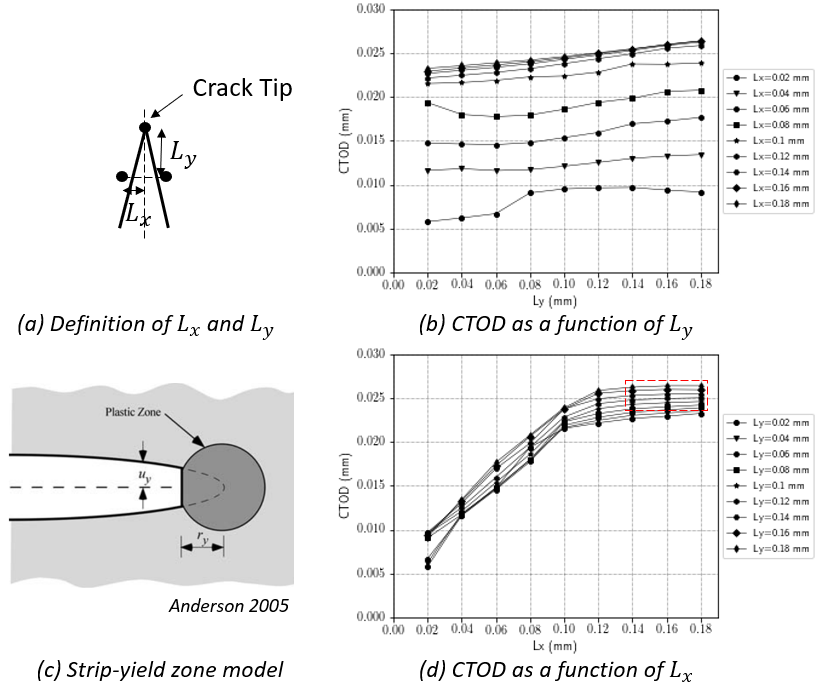}
    \caption{Effect of reference point location on CTOD measurement.}
    \label{fig:ctod}
\end{figure}

Second, the CTOD can be determined by defining a pair of reference points once the crack tip is located. The locations of the reference points have a significant impact on the measured CTOD. As shown in Fig.\ref{fig:ctod}\emph{(a)}, $L_x$ is the distance between the reference point and the crack tip in the direction perpendicular to the crack plane. $L_y$ refers to the distance in the crack plane direction. Fig.\ref{fig:ctod}\emph{(b)} plots the CTOD as a function of $L_y$ for various $L_x$. Fig.\ref{fig:ctod}\emph{(d)} plots the CTOD as a function of $L_x$ for various $L_y$. Although the measured CTOD increases as $L_x$ increases, it could be observed that for $L_x \geq 0.14mm$ and $L_y \geq 0.08mm$, the CTOD attains a plateau at around 0.025mm. This stable plateau region is the result of rigid body motion. It indicates the end of the strip-yield zone, or the boundary of the region undergoing crack tip deformation, as shown in Fig.\ref{fig:ctod}\emph{(c)} (\citeNP{anderson2005fracture}, \citeNP{vasco2019characterisation}).

To determine $\delta_c$, the above-described CTOD measurement procedure must be applied to an image collected in the pre-peak load stage while being close to the peak load point. It is worth noting that a minimum spatial resolution of 1/3 of $\delta_c$ per pixel is recommended to provide accurate crack detection results. For example, a spatial resolution finer than 8 $\mu$m/pixel is recommended in the above case. In other words, a camera with a resolution higher than $5000\times5000$ is advised to detect cracks in a $40mm\times40mm$ region. 

\subsection{Validation}
\subsubsection{Extended Finite Element Model}
The accuracy of an image-based crack detector is often evaluated by comparing it with the ground truth crack pattern that relies on visual recognition \cite{gehri2020automated}. However, small cracks in the early stages of AC fracture tests are critical but often difficult to visualize. This poses a challenge to validate the proposed method experimentally because it is difficult to accurately recognize the ground truth crack patterns. Thus, an XFEM model was developed to validate the proposed DIC-based crack detection method. If the proposed method is valid, given an accurate displacement field from XFEM, it is safe to conclude that the method would work when the DIC-measured displacement field is accurate, which was a challenge solved in the first part this paper.

\begin{figure}[ht!]
    \centering
    \includegraphics[trim=0 0 0 0,clip,width=1\textwidth]{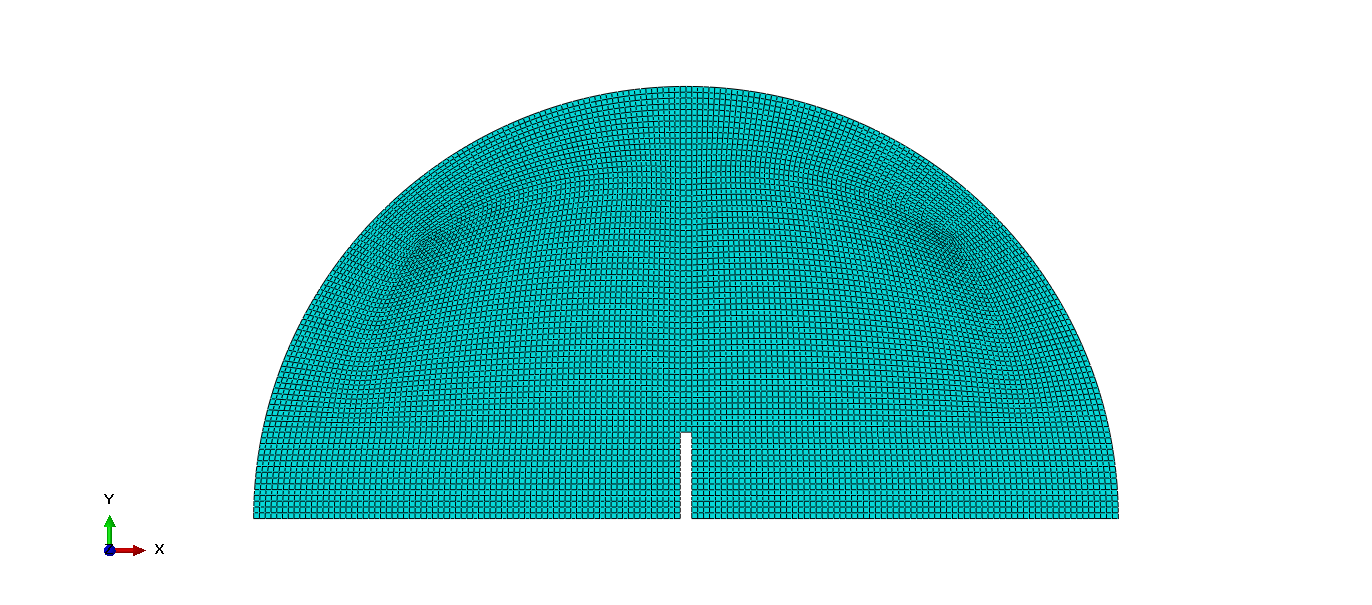}
    \caption{ABAQUS finite element model of the I-FIT.}
    \label{fig:model}
\end{figure}

The XFEM can model crack growth along an arbitrary and solution-dependent path because there is no need for remeshing as a crack grows \cite{belytschko2009review}. Commercial software ABAQUS was used for modeling the I-FIT test. Modeling methods that were validated in previous research were followed (\citeNP{al2015testing}, \citeNP{mahmoud2014extended}, \citeNP{hernandez2018micromechanical}). Plain strain finite element was used to simplify the sample in the 2-D space. Fig.\ref{fig:model} shows the 2-D I-FIT model, which had a diameter of 150mm and a 2-mm-wide and 15-mm-long notch. The geometry used in the model was measured from the test specimen.
Plain strain assumption was used with a thickness of 50mm. The load was applied in displacement-controlled mode at 50mm/min. The specimen was supported by two circular frictionless rods. They were placed symmetrically with respect to the center of the notch. The rods and the loading frame were modeled as rigid analytical surfaces. Their displacements in the x- and y-directions and rotations with respect to the z-axis of the reference nodes were constrained. A temperature field of 25$^{\circ}$C was applied. A quadrilateral element was used in the model. The AC was modeled as a bulk viscoelastic material based on a Prony series expansion of the dimensionless relaxation modulus in ABAQUS \cite{abaqus20146}. The Prony series coefficients reported by \citeN{al2015testing} were used. They were obtained by conducting the complex modulus on a typical Illinois N90 mix following AASHTO TP79. It should be noted that this was the same mix used in the previous section. The $\delta_c$ was experimentally determined to be around 0.025mm, as shown in Fig.\ref{fig:ctod}.

The traction-separation law was used to model damage initiation and evolution. A crack initiates when the principal stress is larger than the maximum allowable. The energy criterion was used to define the damage evolution once the initiation criterion was reached. The model was calibrated by iteratively changing the traction-separation law parameters until the FE model results fit the experimental load-displacement curve. A similar approach was successfully implemented by other researchers (\citeNP{mahmoud2014extended}, \citeNP{lancaster2013extended}).

\begin{figure}[ht!]
    \centering
    \includegraphics[trim=0 0 0 0,clip,width=1\textwidth]{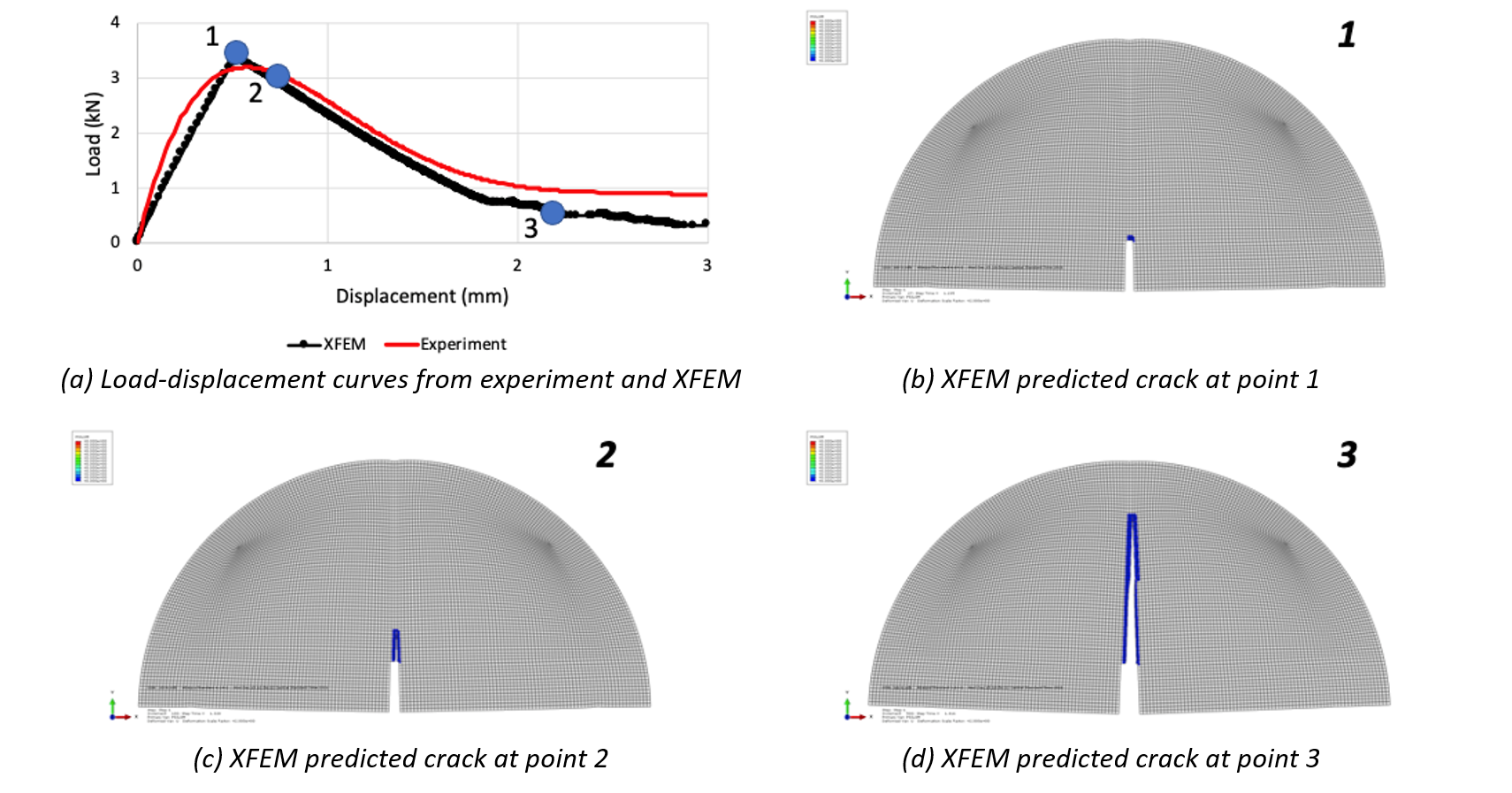}
    \caption{Load-displacement curve and crack propagation predicted by XFEM.}
    \label{fig:xfem_result}
\end{figure}

Fig.\ref{fig:xfem_result} shows the load-displacement curve along with cracks predicted by XFEM. The load-displacement curve obtained from the laboratory test is also shown. The XFEM predicted curve matched well with the experimental results. It is worth noting that, unlike the experiment curve, load and displacement were linearly correlated at the pre-peak-load stage in XFEM. This was because the available traction-separation model in ABAQUS assumes initially linear elastic behavior before the initiation and evolution of the damage \cite{abaqus20146}. Moreover, it could be observed that macro-crack initiated at the notch tip and started to propagate near peak load, which aligned with findings reported by other researchers (\citeNP{al2015testing}, \citeNP{doll2017damage}). These results indicate that the developed XFEM model accurately predicted the crack pattern for the I-FIT.

\subsubsection{Crack Detection Based On Displacement Field}
This section implements the crack detection method described above, given the displacement field from XFEM. The detected cracks were compared with the ground truth crack pattern, a direct output of the XFEM analysis.

First, the $\delta_c$ was determined following the method described in the previous section. Given the quadratic element, size was around 1mm in the XFEM model, and it was observed that the CTOD attained a plateau at approximately 0.023mm for $L_x \geq 0.5mm$ and $L_y \geq 1mm$. The $\delta_c$ obtained from the XFEM simulation was close to this specific mix's experimentally determined value (0.025mm).

\begin{figure}[ht!]
    \centering
    \includegraphics[trim=0 0 0 0,clip,width=1\textwidth]{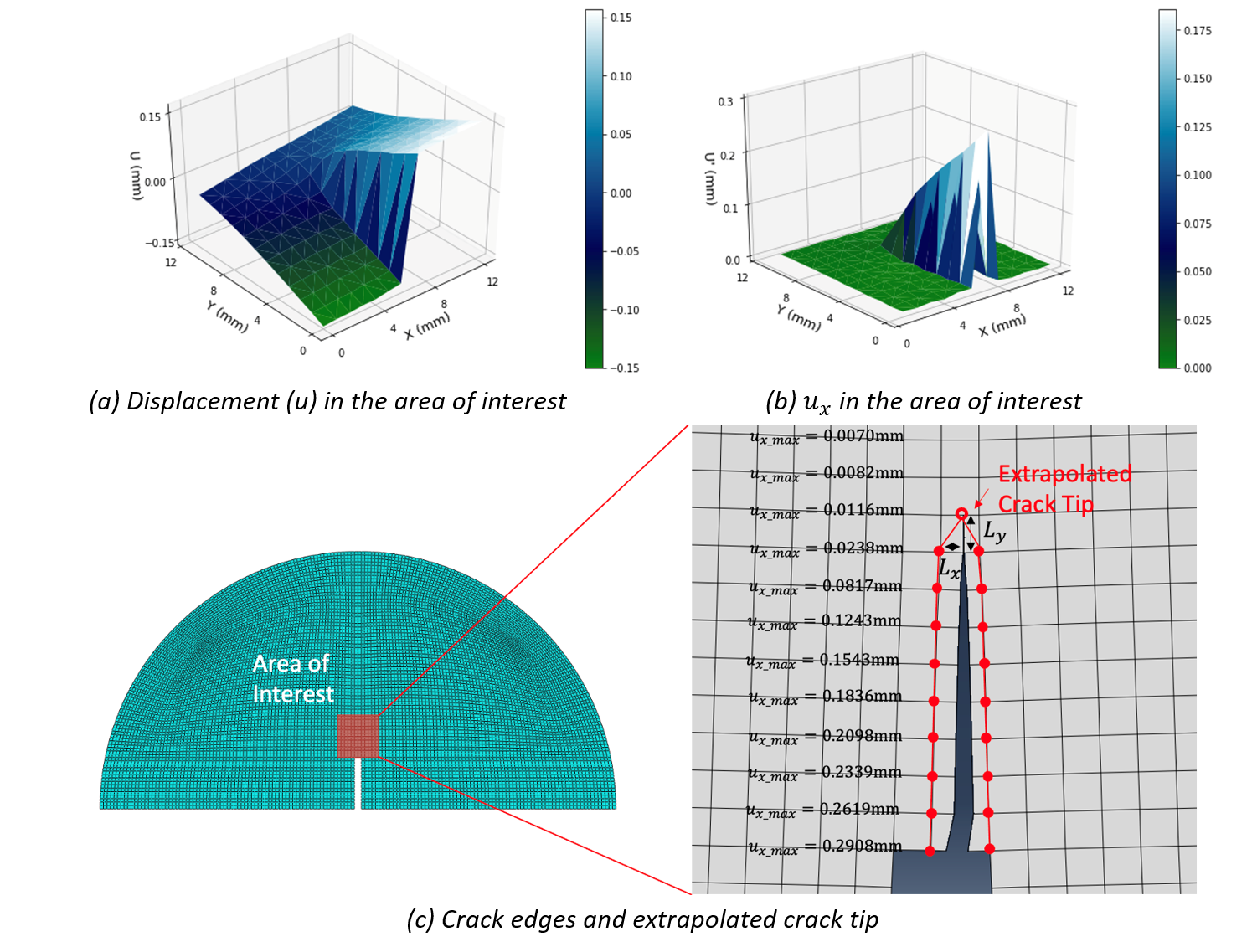}
    \caption{Implementation of the proposed method to detect cracks based on displacement field.}
    \label{fig:crack_result}
\end{figure}

Second, the relative displacement $\mathbf{u}_x$ between neighboring nodes was calculated by filtering $\mathbf{u}$ with a $[-1,1]$ kernel. Fig.\ref{fig:crack_result}\emph{(a)} shows the displacement field $\mathbf{u}$ in the area of interest; Fig.\ref{fig:crack_result}\emph{(b)} plots $\mathbf{u}_x$. 

Third, because a $u_{x_i} \geq \delta_c$ indicates discontinuity, crack edges were labeled by locating the corresponding nodes in the deformed image, as shown in Fig.\ref{fig:crack_result}\emph{(c)}. The crack tip was located by treating the corresponding topmost nodes as the reference points and applying the triangular principle as illustrated in Fig.\ref{fig:ctod}\emph{(a)}. 

It could be observed that the detected crack matches well with the ground-truth crack pattern, which was predicted by the XFEM analysis. It should be noted that the detection accuracy was constrained by the element size used in the XFEM model. The crack edges will be closer to the ground truth if the mesh is more refined. The results indicate that the proposed method accurately characterizes the crack pattern given an accurate displacement field. 

\section{Case Study}
The proposed method could be applied to characterize AC cracking phenomenon, evaluate its fracture characteristics, assess AC mixture testing protocols, and develop theoretical models. In the following case study, the proposed method was applied to measure the crack propagation rate while conducting the I-FIT test, a commonly used fracture test to characterize the cracking potential of an AC mixture. 

\subsection{Test Setup}
As shown in Fig.\ref{fig:dic_set_up}, raw images were collected while conducting the I-FIT. All experiments were conducted at 25\textdegree{}C with a loading rate of 50 mm/min. For each specimen, a random black pattern was applied on top of a layer of white paint. The speckles were sprayed using a fine airbrush. A CCD camera (an Allied Vision Prosilica GX6600 ($6576 \times 4384$ pixels, four fps) with a Tokina AT-X Pro Macro 100 2.8D lens) was positioned perpendicularly to the surface of the I-FIT specimen to collect images during the test. The high-resolution camera captured damage zone evolution in heterogeneous materials such as AC \cite{doll2017damage}. The spatial resolution was around 8 $\mu$m/pixel.

\begin{figure}[ht!]
    \centering
    \includegraphics[trim=0 0 0 0,clip,width=1\textwidth]{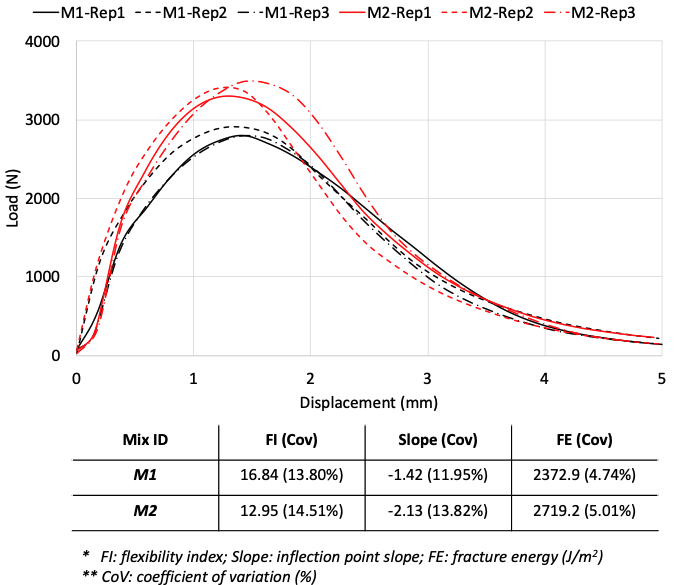}
    \caption{Load-displacement curves and I-FIT results.}
    \label{fig:ifit}
\end{figure}

This case study tested two lab-produced AC mixes, and their design details are summarized in Table \ref{table:mix_design}. The only difference between the two mixes was the binder source, which is known to have a significant impact on AC cracking potential, although they had the same PG \cite{zhu2020quantification}. Three replicates were tested for each mix. Fig.\ref{fig:ifit} illustrates the load-displacement curves and I-FIT results. The lower FI suggested that M2 was more prone to cracking than M1. It could be observed that M2 had a higher peak load than M1, while its post-peak slope was steeper, which indicated faster crack growth during testing \cite{zhu2019influence}. 

\begin{table}
\caption{Mix Design of Two Lab-Produced AC Mixes}
\label{table:mix_design}
\centering
\small
\renewcommand{\arraystretch}{1.25}
\begin{tabular}{l l l l l l l}
\hline\hline
\multicolumn{1}{c}{Mix ID} & \multicolumn{1}{c}{N Design} & \multicolumn{1}{c}{NMAS} & \multicolumn{1}{c}{VMA} & \multicolumn{1}{c}{AC \%} & \multicolumn{1}{c}{Binder Type} & \multicolumn{1}{c}{ABR \%} \\
\hline
M1 & 70 & 9.5mm & 15.2 & 6.4 & PG-64-22-A & 0 \\
M2 & 70 & 9.5mm & 15.2 & 6.4 & PG-64-22-B & 0  \\
\hline\hline
\end{tabular}
\normalsize
\end{table}

\subsection{Displacement Measurement Using 2D-DIC Analysis}
First, the quality of each speckle pattern was evaluated using the mean intensity gradient method (Eq.\ref{eqn:mig}). For M1, replicates one, two, and three have mean intensity gradients of 27.7, 26.1, and 27.4, respectively. For M2, replicates one, two, and three have mean intensity gradients of 23.2, 24.6, and 25.5, respectively. All speckle patterns have good quality because of their large enough mean intensity gradient. 

Second, the subset size was carefully selected as it significantly impacts the accuracy of the measured displacement field \cite{pan2009reliability}. A subset size of $23 \times 23$ was chosen through an iterative process. It provided an adequate spatial resolution to resolve the strain distribution in both asphalt mastics and aggregate particles. As for the step size, although the standard practice is to set it in the range of 1/2 to 1/3 of the subset size for standard applications, for crack measurement, it is suggested to put it to less than 1/6 of the subset size \cite{gehri2020automated}. A step size of 2 was used here. 

Third, multi-seed incremental RG-DIC analysis was conducted to measure the displacement field. The analysis was implemented in NCorr, an open-source software \cite{blaber2015ncorr}.  

\subsection{Crack Measurement}
\subsubsection{Determine Critical CTOD}
The first step in the proposed displacement-field-based crack detection framework is to determine $\delta_c$. 

The crack tip was located by analyzing the image collected in the pre-peak load stage while close to the peak-load point. The method described in the previous section was followed.

\begin{figure}[ht!]
    \centering
    \includegraphics[trim=0 0 0 0,clip,width=0.95\textwidth]{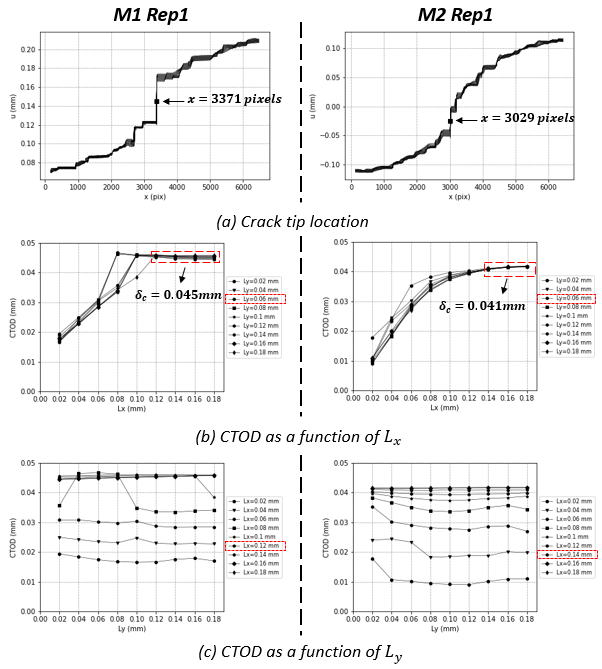}
    \caption{Determination of $\delta_c$.}
    \label{fig:ctod_case}
\end{figure}

The $\delta_c$ was determined by plotting the CTOD as a function of $L_x$ for various $L_y$ and plotting it as a function of $L_y$ for various $L_x$. Fig.\ref{fig:ctod_case} shows the plots for one of the replicates of the two mixes. For M1, it could be observed that for $L_x \geq 0.12mm$ and $L_y \geq 0.06mm$, the CTOD attains a plateau at around 0.045mm. For M2, it could be observed that for $L_x \geq 0.14mm$ and $L_y \geq 0.06mm$, the CTOD attains a plateau at around 0.041mm.

\subsubsection{Crack Detection}
\begin{figure}[ht!]
    \centering
    \includegraphics[trim=0 0 0 0,clip,width=0.95\textwidth]{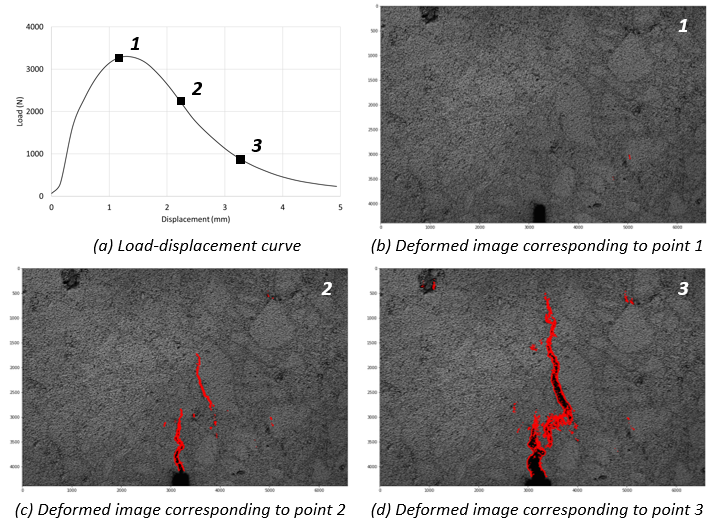}
    \caption{A sequence of deformed images with cracks located.}
    \label{fig:crack_case}
\end{figure}

Given a displacement field ($\mathbf{u}$), the relative displacement ($\mathbf{u}_x$) between neighboring correlation points was calculated by filtering $\mathbf{u}$ by a $[-1,1]$ kernel. A $u_{x_i}$ that is greater or equal to a critical CTOD value ($\delta_c$) means the onset of cleavage fracture. Crack edges were labeled by locating the corresponding correlation points in the deformed image. The crack tip was then found by treating the topmost connected nodes as the reference points and applying the triangular principle as illustrated in Fig.\ref{fig:ctod}\emph{(a)}. Fig.\ref{fig:crack_case} demonstrates a sequence of deformed images (M2 Rep1) with their crack edges highlighted in red. 

\subsubsection{Crack Propagation Analysis}
Crack-propagation speed is one of the primary AC cracking characteristic factors. Most state-of-art AC cracking potential prediction indices rely on an approximate crack-propagation rate because they lack an efficient and reliable method to determine the actual crack-propagation speed. For example, the FI from the I-FIT uses the post-peak inflection-point slope from the load-displacement curve to proxy the crack-propagation rate \cite{al2015testing}. The actual crack-propagation speed could be easily derived using the proposed method.

\begin{figure}[ht!]
    \centering
    \includegraphics[trim=0 0 0 0,clip,width=0.6\textwidth]{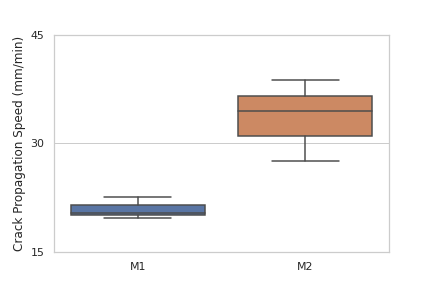}
    \caption{Measured crack propagation speed.}
    \label{fig:speed}
\end{figure}

Fig.\ref{fig:speed} shows the actual mean crack-propagation speed of M1 and M2. The rate was calculated by tracking the crack tip. The mean crack-propagation speed measured on M2 specimens was 60\% faster than on M1. It is worth noting that because of the inhomogeneity of AC, the variability of crack propagation speed is non-negligible. The FI captured the AC material-inherent variability.

\section{Summary}
This paper proposes a framework to accurately detect surface cracks of AC specimens using 2D-DIC. Two challenges that exist in previous research were addressed.

First, multi-seed incremental RG-DIC analysis was proposed to solve the decorrelation issue due to large deformation and discontinuities. The method was validated using six series of synthetic images. It was found that a properly implemented multi-seed incremental RG-DIC analysis could consistently achieve high accuracy up to a strain level of 450\%, even with significant discontinuities (cracks) present in the deformed image. Moreover, a few irregularities and holes commonly seen on an AC specimen surface have no significant impact on measurement accuracy.

Second, existing DIC-based crack detection methods rely on either empirical thresholds or unproven assumptions; a more robust approach was developed to detect cracks using DIC-measured displacement fields. The proposed method uses a $u_{x_i} \geq \delta_c$ or a $v_{y_j} \geq \delta_c$ to define the onset of cleavage fracture. This method relies on the well-developed CTOD concept. The proposed threshold $\delta_c$ has a physical meaning and can be easily determined from DIC measured displacement field. The method accurately characterized the crack pattern using an XFEM model simulating an I-FIT. 

\begin{figure}[ht!]
    \centering
    \includegraphics[trim=0 0 0 0,clip,width=0.45\textwidth]{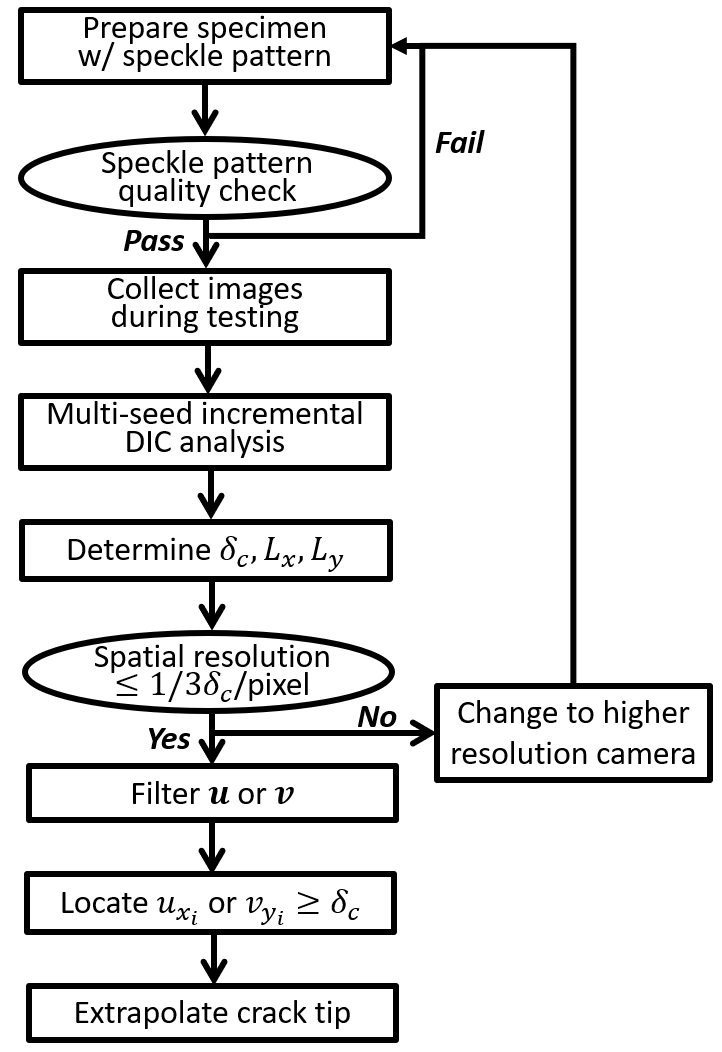}
    \caption{Flowchart of the proposed method.}
    \label{fig:flowchart}
\end{figure}

Fig.\ref{fig:flowchart} summarizes the framework in a flowchart. 

The proposed method could be applied to characterize AC cracking phenomenon, evaluate its fracture characteristics, assess AC mixture testing protocols, and develop theoretical models. A case study was conducted on two AC mixes to compare their crack propagation speeds. The measured rates successfully distinguished the crack potential of the two mixes.

\section{Limitations and Recommendations}
The followings are the limitations of this study and relative
suggestions for future research:
\begin{itemize}
    \item The incremental DIC analysis suffers from a gradual accumulation of errors. To further increase the accuracy of DIC analysis under large deformation, it is suggested to explore other approaches such as scale-invariant feature transformation (SIFT)-aided DIC or quasi-conformal mapping (\citeNP{yang2020sift}, \citeNP{ye2022digital}).
    \item The framework proposed in this paper is only valid for fracture tests. Developing a crack detection method for strength tests like the indirect tension asphalt cracking test (IDEAL-CT) is recommended.
    \item This paper only investigated 2-D DIC, which requires the optical axis of a camera to be placed perpendicular to the specimen surface. It is worth evaluating the proposed framework using 3-D DIC.
\end{itemize}

\section{Data Availability Statement}
All data, models, or code that support the findings of this study are available from the corresponding author upon reasonable request.

\section{Acknowledgment}
The authors would like to thank the research engineers and students at the Illinois Center for Transportation for their input and support during this study. Special
thanks to Greg Renshaw, Uthman Mohamed Ali, and Jose Julian Rivera Perez. The contents of this paper reflect the view of the authors, who are responsible for the facts and the accuracy of the data presented here. 

%
%
\bibliography{ascexmpl-new}

\end{document}